%% file: root.tex
\documentclass[letterpaper, 10 pt, conference]{ieeeconf}  

\IEEEoverridecommandlockouts                              

\usepackage{todonotes}
\usepackage{graphics} 
\usepackage{graphicx} 
\usepackage{amsmath} 
\usepackage{amsfonts}
\usepackage{amssymb}  
\usepackage{bm}
\usepackage{multirow}
\usepackage[keeplastbox]{flushend}

\usepackage{color}
\definecolor{morado}{rgb}{1,0,1}
\definecolor{gris}{cmyk}{0,0,0,0.1}
\definecolor{amarillo}{cmyk}{0,0,0.6,0}
\definecolor{blanco}{cmyk}{0,0,0,0}
\definecolor{negro}{cmyk}{1,1,1,0}
\definecolor{orange}{cmyk}{0,0.5,0.8,0}
\definecolor{verde}{cmyk}{0.7,0.1,1,0}

\title{\LARGE \bf
Deconvolutional Networks for Point-Cloud Vehicle Detection\\
and Tracking in Driving Scenarios 
}

\author{V\'ictor Vaquero$^*$, Ivan del Pino$^*$, Francesc Moreno-Noguer, Joan Sol\`a, Alberto Sanfeliu and Juan Andrade-Cetto
\thanks{*These authors contributed equally to this work. {\tt \small \{vvaquero,idelpino\}@iri.upc.edu}.}
\thanks{The authors are with the Institut de Rob\`otica i Inform\`atica
Industrial, CSIC-UPC, Llorens Artigas 4-6, 08028 Barcelona, Spain.}%
\thanks{This work has been supported by the 
Spanish Ministry of Economy and Competitiveness projects ROBINSTRUCT (TIN2014-58178-R) and COLROBTRANSP (DPI2016-78957-R), 
by the Spanish Ministry of Education FPU grant (FPU15/04446), 
the Spanish State Research Agency through the Mar\'ia de Maeztu Seal of Excellence to IRI (MDM-2016-0656) 
and by the EU H2020 project LOGIMATIC (H2020-Galileo-2015-1-687534).
The authors also thank Nvidia for hardware donation under the GPU grant program.}%
}

\begin{document}

\maketitle
\thispagestyle{empty}
\pagestyle{empty}

\begin{abstract}
Vehicle detection and tracking is a core ingredient for developing autonomous driving applications in urban scenarios. Recent image-based Deep Learning (DL) techniques are obtaining breakthrough results in these perceptive tasks. 
However, DL research has not yet advanced much towards processing 3D point clouds from lidar range-finders. These sensors are very common in autonomous vehicles since, despite not providing as semantically rich information as images, their performance is more robust under harsh weather conditions than vision sensors. 
In this paper we present a full vehicle detection and tracking system that works with 3D lidar information only. 
Our detection step uses a Convolutional Neural Network (CNN) that receives as input a featured representation of the 3D information provided by a Velodyne HDL-64 sensor and returns a per-point classification of whether it belongs to a vehicle or not. 
The classified point cloud is then geometrically processed to generate observations for a multi-object tracking system implemented via a number of Multi-Hypothesis Extended Kalman Filters (MH-EKF) that estimate the position and velocity of the surrounding vehicles. 
The system is thoroughly evaluated on the KITTI tracking dataset, and we show the performance boost provided by our CNN-based vehicle detector over a standard geometric approach. Our lidar-based approach uses about a $4\%$ of the data needed for an image-based detector with similarly competitive results.
\end{abstract}

\input{1_2_Intro_Related}

\input{3_Method}

\input{4_Experiments}
\input{5_Conclusions_Future}
\bibliographystyle{./IEEEtran}
\bibliography{./IEEEabrv,./references}

\end{document}

%% file: 1_2_Intro_Related.tex
\section{INTRODUCTION}

Autonomous driving (AD) is nowadays a reality. The main reasons for  this success are twofold. On the one hand, research advances in related areas such as machine learning and computer vision are obtaining a high level of scene comprehension of the vehicle surroundings. 
On the other hand, new hardware and on-vehicle sensors are providing the community with enough data to develop new and robust perception algorithms as well as the ability to process them in real time.

However, there is still a long road until fully autonomous vehicles (AV) drive along totally free in our cities. 
Urban traffic is very challenging and dynamic, with numerous intervening elements like pedestrians, cyclists, other vehicles and even street furniture. 
To this end, accurate detection and tracking algorithms are elements of vital importance in AVs. These systems must be robust enough to recognise, understand, and act in response to any possible situation while assuring the safety of drivers, pedestrians, and other elements in our roads.

\begin{figure}[t]
\centering
 \centerline{
  	\includegraphics[width=0.95\columnwidth]{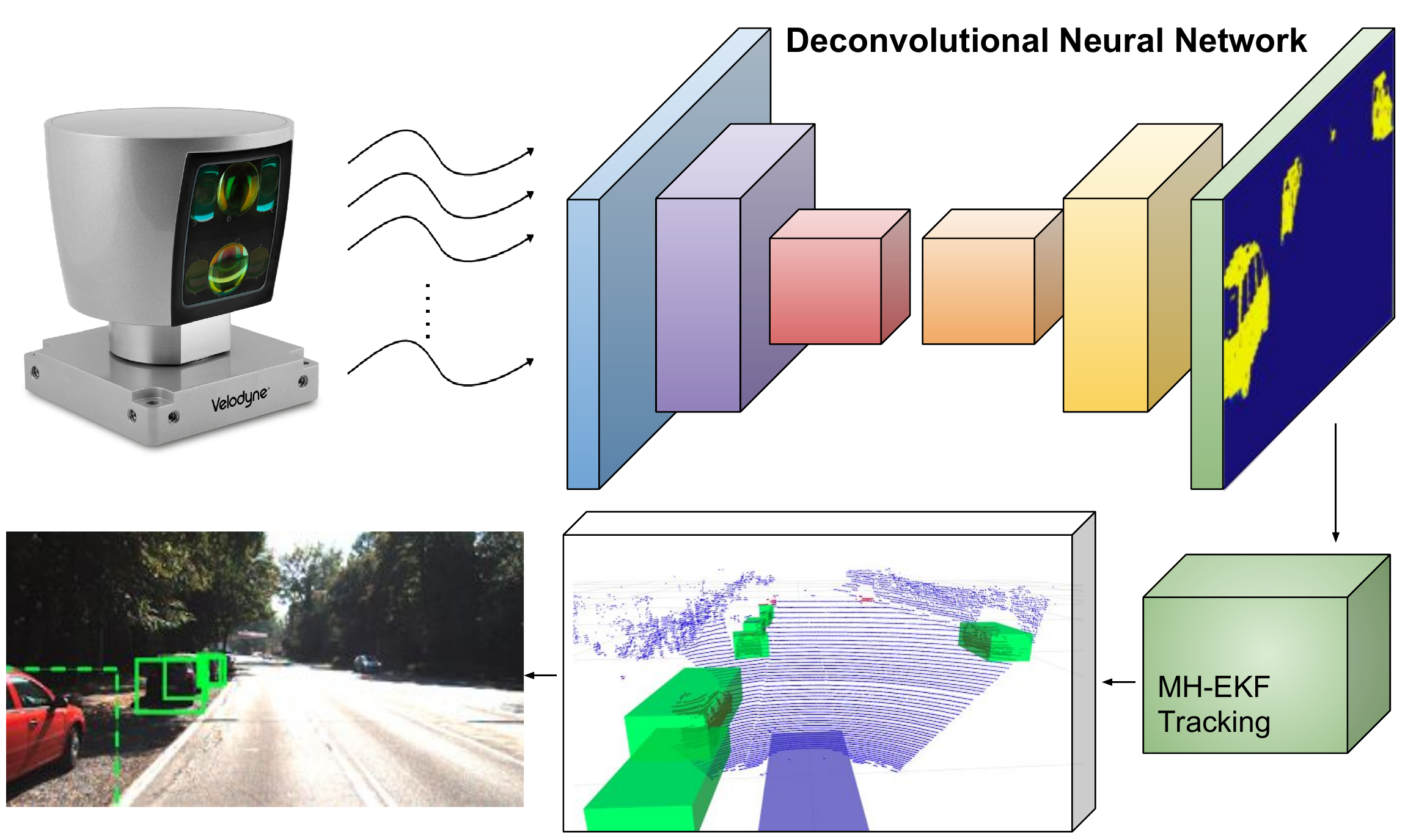}}
  
  \caption{We introduce a novel CNN-based vehicle detector on 3D range data. The proposed model is fed with an encoded representation of the point cloud and computes for 3D each point its probability of belonging to a vehicle. 
  The classified points are then clustered generating trustworthy observations that are fed to our MH-EKF based tracker. Note: Bottom left RGB image is shown here only for visualization purposes.
  \vspace{-3mm}} 
  \label{fig:init}
\end{figure}

With the advent of deep learning technologies, image-based scene understanding involving tasks such as object detection, semantic segmentation or motion capture have experienced an impressive performance and accuracy boost \cite{krizhevsky2012imagenet, ren2015faster, long2015fully, vvaquero2017flow}. However, image-based methods may suffer a high performance decrease in real driving scenarios under harsh environmental conditions e.g. heavy rain, snow, fog, or even night scenes.
To avoid such situations and increase robustness, redundancy must be included in AVs. This is commonly tackled by creating autonomous perception systems that rely on other sensors such as radar or 3D lidar range-scanners. 

Lidar sensors are specially suitable for AD purposes since they provide very accurate spatial information of the environment, are robust to hard climate conditions and their performance is almost independent on the illumination of the scene. 
Yet, deep learning methods deployed over 3D-lidar point clouds are far from the successful performance achieved on 2D-RGB images. This is mainly due to the computational burden introduced by the change in problem dimensionality as well as lack of annotated training data.

We present a robust and accurate vehicle detection and tracking system that uses solely 3D lidar information as input. A sketch of the developed system is shown in Fig.~\ref{fig:init}.
The main core of the presented approach is a tailored Fully Convolutional Network (FCN)~\cite{long2015fully} trained to detect vehicles from featured range and reflectivity representations of the 3D point cloud provided by a Velodyne 64-HDL sensor. 
Our FCN fulfils this task by performing a point-wise classification of whether each 3D point belongs to a vehicle or not. 
Positive samples are then clustered and 2D vehicle poses are obtained. This is performed by choosing the best fitting oriented bounding boxes $(x,y,\theta)$ over the external perimeter resulting after projecting the clusters to the ground plane.
This 2D information is finally fed to a tracker based on a Multi-Hypothesis Extended Kalman Filter (MH-EKF) which, along with extracted 3D features such as the heigh of the corresponding cluster, provide the final results on 3D tracking. 

We test our system over the Kitti tracking benchmark~\cite{Geiger2012CVPR}, where lidar-only methods are heavily penalised due to the image-based 2D evaluation measurements. However, we show the competitiveness of our approach and validate the hypothesis of using CNN-based lidar detectors against other geometrical methods.


 \section{RELATED WORK}

Deep learning techniques, and more specifically Convolutional Neural Networks (CNNs), have demonstrated an outstanding performance in classical computer vision problems such as object classification \cite{krizhevsky2012imagenet, he2016residual}, detection \cite{ren2015faster, sermanet2014overfeat}, and semantic segmentation \cite{long2015fully}. However, CNNs have not yet deployed its potential over range lidar point clouds. 
We next review some approaches proposed to detect objects in such 3D sparse point clouds, and how Deep Learning methods are approaching this task.

\vspace{1mm}
\noindent\textbf{Classical Object Detection in Lidar Point Clouds.}
Extensive literature exists about detecting objects in lidar-generated point clouds. Most common, are segmentation approaches that try to cluster closer points together and classify the resulting groups \cite{arras2007using, vaquero2016low, douillard2011segmentation, mertz2013moving}. These methods typically hold for both single (2D) and multi-layer (3D) lidars. For the latter, voting schemes are also used to vertically fuse single-layer clusters, obtaining part-based models of the objects \cite{mozos2010multi, spinello2010layered}. 
In autonomous driving applications, segmentation approaches previously tend to remove the ground-plane \cite{petrovskaya2009model, teichman2011towards}, easing the clustering and classification steps. This heuristic is useful for the computation of the bounding box of the detected object, as will be shown in Section \ref{sec:bbs}.
Other subtle clustering methods create graphs over pre-processed 3D voxels, exploiting their connectivity later in the classification step \cite{wang2012could, triebel2010segmentation, papon2013voxel}.

Recent methods scrutinize directly the 3D range scan space with sliding window approaches. 
\textit{Vote3D}~\cite{wang2015voting} for example encode the sparse lidar point cloud in a grid with different features such as mean and variance of intensity, a grade of occupancy, and other three different shape factors. The resulting representation is scanned in a sliding manner with 3D windows of different sizes and orientations, classifying the final candidates using SVMs and a voting scheme. 

For the classification of point cloud clusters, the standard approach inherited from RGB algorithms, is to hand-craft features such as spin images, shape models or geometric statistics. Details of the most commonly used 3D features can be found in \cite{behley2012performance}. 
Learning procedures have also been used to obtain useful features via sparse coding, such as the work in \cite{de2013unsupervised}.

\vspace{1mm}
\noindent\textbf{Deep Learning for 3D Lidar Object Detection.}
Following the feature learning tendency, and aware of the success of CNN models, a few authors are applying convolutions over 3D lidar point clouds. 
For example, 3D convolutions, which are commonly used for video analysis (devoting its third dimension to the time variable) have been applied for 3D vehicle detection in \cite{li20163d}. However, due to the high dimensionality and sparsity of the data, deploying this methods over point clouds implies a high computational burden, which is not yet practical for on-line applications.
Reformulating convolutions is a solution. In this way, \textit{Vote3D} has been very recently extended in \cite{engelcke2017vote3deep} by replacing SVMs with novel sparse 3D convolutions that act as voting weights for predicting the detection scores. Other methods design and apply sparse convolutions, such as \cite{jampani2016learning, graham2015sparse}.

Another adopted approach is to obtain equivalent 2D representations of the 3D point cloud to apply the well know and optimized 2D convolution tools. In this way, \cite{li2016vehicle} built a front view representation in which each element encodes a ground-measured distance and height of the 3D point. On top of this representation they apply a Fully Convolutional Network trained to predict the objectness of each point, and simultaneously perform a regression of the 3D bounding box of each vehicle. 
Similarly, we classify 3D points as belonging to vehicles or background although our image-like lidar representation includes direct information about range and reflectivity of the points and we use a more advanced deconvolutional architecture, as it will be shown in Sections \ref{sec:2d_data} and \ref{sec:net}.

The very recent evolution of \cite{li2016vehicle} combines their front view representation of the lidar information with a bird's eye view to generate accurate 3D bounding box proposals \cite{chen2016multi}. These are later fused with RGB images in a region-based fusion network, obtaining state of the art results in the detection challenge of the Kitti dataset. However, this method does not fulfil the lidar-only requirement that we impose in our work.

%% file: 3_Method.tex
\section{VEHICLE DETECTION \& TRACKING}

We reformulate the task of detecting vehicles in lidar point clouds as a per-point classification problem in which we want to obtain the probability of each sample to be a vehicle, therefore:
$p(k|p_i)$, where $k \in \{$vehicle, no-vehicle$\}$; $i~\in~1,...N$, and $p_i \in \mathbb{R}^3$ represents each point of the point cloud $\mathcal{P}$ in the Euclidean space. 

\subsection{2D Representation of Range Data}
\label{sec:2d_data}
\begin{figure}[t]
\centering
 \centerline{
  	\includegraphics[width=0.98    \columnwidth]{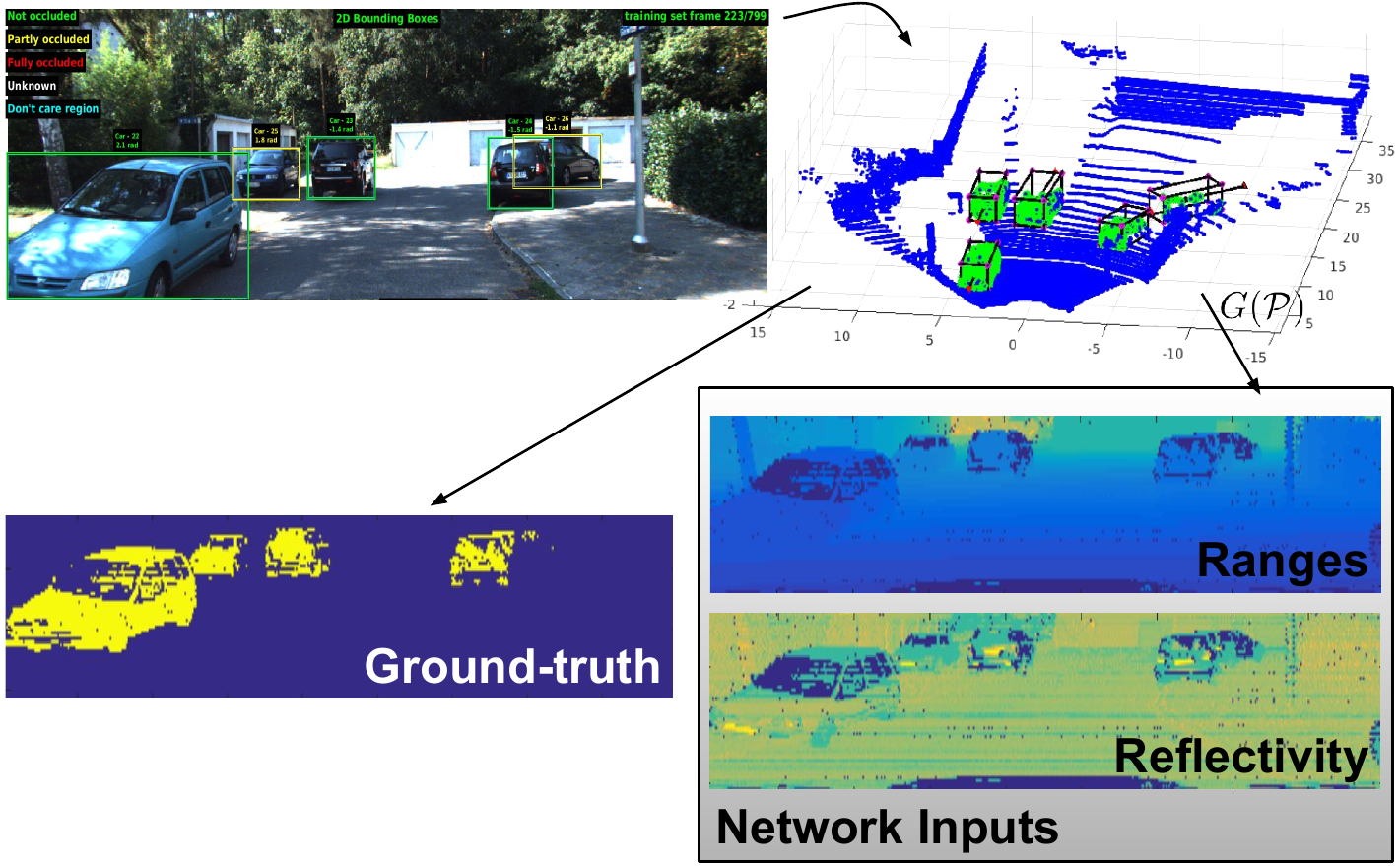}}
  
  \caption{To obtain a useful input for our CNN-based vehicle detector, we project the 3D point cloud raw data to a featured image-like representation containing ranges and reflectivity information by means of $\mathbb{G}(\cdot)$. 
  Ground-truth for learning the proposed classification task is obtained by first projecting the image-based Kitti tracklets over the 3D Velodyne information, and then applying again $\mathbb{G}(\cdot)$ over the selected points.}
  
  \label{fig:imdb}
\end{figure}

To efficiently exploit the successful deep convolutional architectures, we project our point cloud to an image-like representation through $G(\mathcal{P}) \in \mathbb{R}^{H \times W}$. This process is sketched in Fig.~\ref{fig:imdb}.

To obtain the transformation $G(\cdot)$, we first project the 3D Cartesian point cloud to spherical coordinates $sph(p_i)~=~\{\phi_i,\theta_i, \rho_i\}$. According to the Velodyne HDL-64 specifications, elevation angles $\theta$ are represented as a $H \in \mathbb{R}^{64}$ vector with a resolution $\Delta\theta$ of $1/3$ degrees for the upper laser rays and $1/2$ degrees the lower half respectively. Moreover, $G(\cdot)$ needs to restrict the azimuth field of view, $\phi \in [-40.5,40.5]$ to avoid the presence of unlabelled vehicle points, as the Kitti tracking benchmark has labels only for the front camera viewed elements. The azimuth resolution was set to a value of  $\Delta\phi=0.18$ degrees according to the manufacturer, and hence lying in $W \in \mathbb{R}^{451}$. 
Each $H,W$ pair encodes the range ($\rho$) and reflectivity of each projected point, so finally our input data representation lies in an image-like space $G(\mathcal{P}) \in \mathbb{R}^{64 \times 451 \times 2}$. 

To get the equivalent ground-truth representation needed for the learning process, we use the Kitti tracklets, which are given in the RGB space. These tracklets are converted to 3D and $\mathcal{P^*}$ is obtained after labelling the inlier points. These ground-truth 3D class labels are also encoded as one channel in the $G(\cdot)$ image-like space, with pixels taking values of $1$ for background and $2$ for vehicles. Within the `vehicle' class, we consider the associated Kitti classes for \textit{car}, \textit{van} and \textit{truck}. Yet,  the evaluation methods of the benchmark do not have into account \textit{truck} classes, which may penalize our Kitti measured performance.

\subsection{Deconvolutional Networks for Vehicle Detection}
\label{sec:net}
\begin{figure*}[t]
\centering
 \centerline{
  	\includegraphics[width=1  \textwidth]{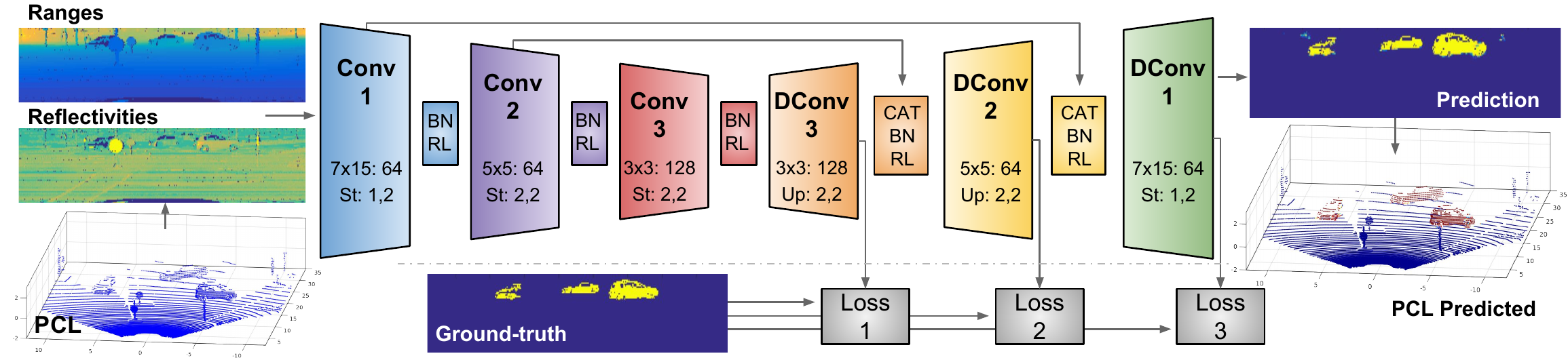}}
  
  \caption{Our network encompasses only convolutional and deconvolutional blocks followed by Batch Normalization (BN) and ReLu (RL) non-linearities. The first three blocks conduct the feature extraction step controlling, according to our vehicle detection objective, the size of the receptive fields and the feature maps generated. The next three deconvolutional blocks expanse the information enabling the point-wise classification. After each deconvolution, feature maps from the lower part of the network are concatenated (CAT) before applying the normalization and non-linearities, providing richer information and better performance. During training, three losses are calculated at different network points, as shown in the bottom part of the graph.}
  \label{fig:net}
\end{figure*}

For the per-pixel vehicle classification task, we propose the deconvolutional architecture shown in Fig.~\ref{fig:net}, having in mind the recent success of these architectures. Here we disclose some of the key insights of our design. 

To reduce the imbalance in the vertical and horizontal dimensions of the Velodyne representations and obtain more tractable intermediate feature maps, in the first convolutional layer we impose twice horizontal than vertical. 
Initial convolutional filter sizes are also designed according to the shape of the vehicles observed in the new representation, so that to obtain a receptive field consistent with it.
Moreover, to address the disproportion between the number of samples of each class, we penalize misclassification of the positive vehicle samples with $\omega$ as seen in Eq.~\ref{eq:2}.

Since our contractive-expansive design could suppose an information bottleneck in the narrow layers, we introduce skip connections concatenating equivalent feature maps from both parts. These links help the learning process of the lower layers by back-propagating purer gradients from the upper parts.
Finally, we state the classification problem at different resolutions of the network in order to obtain a direct and finer control of the learning process. This is done by including intermediate losses that guide the network faster to a correct solution, introducing new valuable gradients at these middle levels.

Hence, the network is trained via end-to-end back-propagation guided by the following loss function:

\begin{equation}%
\label{eq:CaF}
	\mathcal{L}(\hat{\mathcal{Y}}, \mathcal{Y}) =
	\sum_{r=1}^{3} \lambda_{r} \mathcal{L}_{r}(\hat{\mathcal{Y}_r}, {\mathcal{Y}_r}),
\end{equation}

\noindent where $r$ represents the intermediate loss-control positions, $\lambda_{r}$ are regularization weights for the loss at each resolution, and $\hat{\mathcal{Y}}, \mathcal{Y}$ are respectively the predictions and ground-truth classes at those resolutions. 

In our approach, $\mathcal{L}_r$ is a multi-class Weighted Cross Entropy loss (WCE)~\cite{RosArxiv16}, defined as:

\vspace{-5mm}
\begin{equation}
\mathcal{L}_{r}^{WCE} = -\sum_{i,j,k}^{H_r,W_r,K_r} \omega(\mathcal{Y}_{i,j}) \textit{Id}_{[\mathcal{Y}_{i,j}]} \text{log}(\hat{\mathcal{Y}}_{i,j,k}), 
\label{eq:2}
\end{equation}

\noindent where $\textit{Id}_{[x']}(x)$ is an index function that selects the probability associated to the expected ground truth class and $\omega(k)$ is the previously mentioned class-imbalance regularization weight computed from the training set statistics.

\subsection{Obtaining the Vehicle Bounding Boxes}
\label{sec:bbs}

The output $\hat{\mathcal{Y}}$ of the designed network is a matrix laying in the $G(\cdot) \in \mathbb{R}^{64x451}$ space where each pixel represents the probability of the corresponding 3D point to belong to a vehicle. 
In order to obtain the vehicle bounding boxes needed for the tracking and evaluation steps, we first apply the inverse transformation $G^{-1}(\hat{\mathcal{Y}})$ to the network output. In this way, we obtain the equivalent classified 3D point cloud $\mathbb{\hat{P}} \in \mathbb{R}^3$ that is finally clustered by means of an euclidean threshold. 
For each resulting cluster a set of features is extracted to build the EKF observation vector, such as the centroid, height and `vehicleness' (calculated as the mean of the classification scores given by the network to that cluster).
Clusters are then projected over a polar grid on the ground plane, accounting for each azimuth angle only the closest point to the sensor. The aim of this process is to get the external perimeter of each vehicle, which will ease the task of fitting a 2D oriented rectangular model to any of them.

The oriented bounding box fitting process consist on first performing an angular swept of bounding boxes in the $[-\frac{\pi}{4}, \frac{\pi}{4}]$ interval. For each one we cast simulated 2D Velodyne rays to obtain the geometrically equivalent impact over the boxes. The best 2D fitting box is then chosen as the one with the minimum mean square distance between the real vehicle detected points and the simulated ones.
Finally we extract useful features for the vehicle observation vector of the tracker, such as width, length and centre of the 2D bounding box, as well as recover the 3D box using the previously calculated cluster height.

\subsection{Lidar-based Vehicle Tracking}
\label{sec:trackingMethod}

We implemented a multi-hypothesis extended Kalman filter (MH-EKF) for tracking bounding boxes according to a realistic motion model suited for wheeled vehicles in road environments. 
As vehicles transit on the road plane, the 2D bounding boxes (BB) are considered for tracking. 
Since the true vehicle dimension and centroid are not measurable through simple detections, we locate the BB origin at the closest visible corner, which is indeed measurable.
For each BB we start a MH-EKF, which tracks its 2D position, orientation and velocity. 
The state vector is
\begin{equation}
{\bf x} = \begin{bmatrix}
{\bf p}^\top&\theta&v&\rho
\end{bmatrix}^\top
=\begin{bmatrix}
x&y&\theta&v&\rho
\end{bmatrix}^\top
\end{equation}
where ${\bm p}\triangleq(x,y)$ and ${\bm \pi}\triangleq(x,y,\theta)$ are the BB's position and pose, $v$ is the linear velocity in the local $x$ direction, and $\rho$ is the inverse of the curvature radius (so that the angular velocity is $\omega=v\rho$).

Due to the limited geometrical information of the detected bounding boxes, we establish multiple hypotheses for the box motion: i.e, one moves along the main horizontal axis, and another one across it. Initially we assign uniform weights to all hypotheses $w_i=1/N,~ i\in[1,\cdots,N]$. Each EKF estimation ${\bf x}_i$ evolves normally according to the motion model ${\bf x}\gets f({\bf x}, {\bf w}, \Delta t)$, described as
\begin{align}
{\bf p}_i &\gets {\bf p}_i + v_i
\begin{bmatrix}
cos(\theta_i)\\
sin(\theta_i)
\end{bmatrix}
\Delta t \\
\theta_i &\gets \theta_i + v_i\rho_i\Delta t \\
v_i &\gets v_i + w_v \\
\rho_i &\gets \rho_i + w_\rho
~,
\end{align}
where $\gets$ represents a time-update; and the measurement model ${\bf y} = h({\bf x}_i)+{\bf v}$, described as
\begin{align}
{\bf y} &= ({\bm \pi}_i \ominus {\bm \pi}_V) \ominus {\bm \pi}_S + {\bf v}
~.
\end{align}
In these models, ${\bf w}=(w_v,w_\rho)$ and ${\bf v}$ are white Gaussian processes, $\ominus$ is the subtractive frame composition, ${\bm \pi}_V$ is the pose of the own vehicle, which is considered known through simple odometry, and ${\bm \pi}_S$ is the sensor's mounting pose in the vehicle.
The measurement ${\bf y}=(x_{S}, y_{S}, \theta_{S})$ matches the result of the detection algorithm in sensor frame.
At each new observation, the weights are updated according to the current hypothesis likelihood $\lambda_i$, that is,
\begin{align}
\lambda_i &= \exp(-\dfrac12{\bf z}_i^\top{\bf Z}_i^{-1}{\bf z}_i) \\
w_i &\gets w_i \lambda_i,
\end{align}
where ${\bf z}_i={\bf y}-h({\bf x}_i)$ is the current measurement's innovation, and ${\bf Z}_i$ its covariances matrix. Weights are systematically normalized so that $\sum w_i=1$.
Finally, when a weight drops below a threshold $\tau$, its hypothesis is discarded. A few observations after the initial detection only one hypothesis remains for each filter.

This basic scheme is modified with the management of the visible corners: in cases of partial occlusion or vehicle overtake, we may have to switch the initially detected corner (which may has gone out of sight) by the closest currently visible one. This is done by trivially updating the $(x,y)$ states to the new visible corner, leaving all other states untouched.

\pagebreak

%% file: 4_Experiments.tex
\section{RESULTS}

\begin{figure*}[t]
\centering
 \centerline{
  	\includegraphics[width=1  \textwidth , height=14cm]{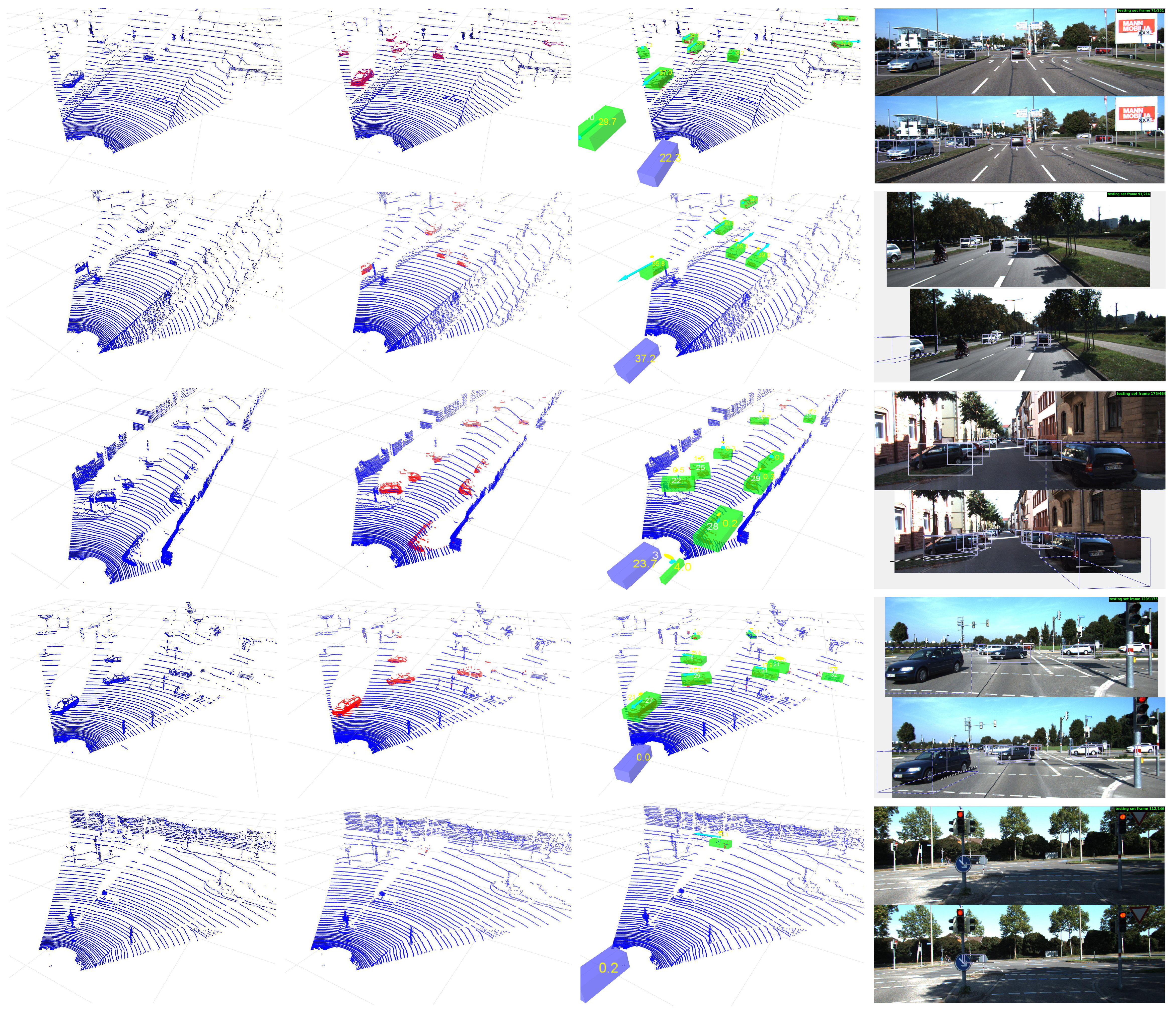}}
  \caption{Qualitative results of our system. All images are taken from the testing set of the Kitti Tracking benchmark, so none of them were previously seen by our Deep Lidar detector. In columns, images show the raw input point cloud, the Deep detector output, the final tracked vehicles and the RGB projected bounding boxes submitted for evaluation. Note how despite the scarce information provided by the lidar, our system is able to detect (red coloured points) and track (green boxes) vehicles in complex urban environments, even when they are partially occluded  (bottom row).}
  \label{fig:res}
\end{figure*}

We measure the performance increase provided by our CNN-based lidar detector over the presented MH-EKF tracker in the Kitti Tracking benchmark. 
Additionally, we provide insights of the precision/recall obtained by our DeepLidar detector and a qualitative evaluation of the full system that can be seen in Fig.~\ref{fig:res}.

The Kitti tracking benchmark is composed of a training set of $8,000$ Velodyne scans grouped into $21$ different sequences covering diverse urban environments. For these, 3D tracklets defined over the corresponding RGB images are provided. In addition, $11,095$ scans are given in a test set grouped into $29$ sequences with no annotations provided. 
Velodyne data timestamps are not given for any of the sequences in the tracking benchmark. As our tracker integrates the observations with the vehicle odometry, we therefore had to create synthetic timestamps simulating the Velodyne data at $10$Hz as specified by the manufactures.

\subsection{Full Working System}

We designed and trained our Deep Learning models using MatConvNet. Networks are initialized with the He's method~\cite{he2015delving} and use Adam optimization with the standard parameters $\beta_1 = 0.9$ and $\beta_2 = 0.999$. 
Data augmentation is done with a $50\%$ chance by performing only horizontal flips in order to preserve the geometry properties of the lidar information.
The training process is performed on a single NVIDIA K40 GPU for $200,000$ iterations with a batch size of $20$ Velodyne scans per iteration. The learning rate is fixed to $10^{-3}$ during the first $150,000$ iterations after which, is halved. We select the imbalance regularizator $\omega$ as $25$ and the loss regularizers $\lambda_r$ as $1$, $0.7$ and $0.5$ respectively.

For the clustering step, we group 3D points imposing a maximum distance of $1m$. After that, clusters with less than $25$ points or with a radio below than $50cms$ are discarded. The remaining clusters and its respective bounding boxes are then converted to ROS format and serve as input observations for our tracker. 

Each detected vehicle is assigned a bi-hypothesis MH-EKF: one hypothesis assumes motion along the longest rectangle dimension; the other across it. Each MH-EKF is set up as summarized in Tab.~\ref{tab:MHEKF}, which shows (in order) 
the number of hypotheses, 
the pruning threshold, 
the initial means and sigmas of each hypothesis, 
and the process and measurement noises' sigmas. The orientation observation noise is set to the maximum possible error for a rectangle, $\frac{\pi}{2}$, and dynamically adjusted by a factor that depends on the model fitting error and the cluster dimensions:  
\begin{equation}
c = k\frac{\sum{(r_p - r_v)^2}}{n(w+l)^2}
\end{equation}
where $r_p$ and $r_v$ are the ranges of the real points and the virtual ones, $n$ is the number of points, $w$ and $l$ are the width and length of the virtual rectangle, and $k$ is a tuning parameter experimentally set to 100. All metric units ($r_p$, $r_v$, $w$, $l$) are expressed in meters.

\begin{table}[tb]
\centering
\caption{Parameter setup for all MH-EKFs}
\label{tab:MHEKF}
\begin{tabular}{c|c|c|}
param & value & units / comment \\
\hline
\hline
$N$ & 2 & along and across\\
$\tau$ & 0.001 &\\
\hline
${\bf x}_1$ & $x_S, y_S, \theta_S, 0, 0$ & m, m, rad, m/s, 1/m\\
$\sigma_1$ & $2, 2, \pi/2, 20, 0.2$ & m, m, rad, m/s, 1/m\\
${\bf x}_2$ & $x_S, y_S, \theta_S + \pi/2, 0, 0$& m, m, rad, m/s, 1/m\\
$\sigma_2$ & $2, 2, \pi/2, 20, 0.2$ & m, m, rad, m/s, 1/m\\
\hline
$\sigma_{\bf w}$ & $ 0.5, 0.01$ & m/s, 1/m \\
$\sigma_{\bf v}$ & $0.9, 0.9, c\,\pi/2 $ &  m, m, rad \\
\hline
\end{tabular}
\end{table}

\subsection{Experiments}

We first evaluate the performance of our point-wise convolutional vehicle detector. 
In order to avoid over-fitting and audit the generalization capacities of the proposed architecture, we perform a 4-fold cross-validation step during training. In this way, we train the same architecture selecting each time different sequences to compose a validation set of around $1,000$ samples in a manner that the \textit{vehicle} vs \textit{non-vehicle} points ratio in the resulting sets remains similar.
Averaged results show that our detector is able to classify Velodyne 3D points from the validation sets with a mean precision of $82.3\%$ and a recall of $87.6\%$. 
Notice that this measures are point-wise, and do not refer to the number of vehicles, but to the mean amount of points correctly classified for each scan. However this results demonstrates the capacity of the trained model to retain generalized information of vehicles according to our input representations, which enables to train the final model with the full $8,000$ samples of the Kitti training set.

In addition, we measure the contributions of our DeepLidar vehicle detector applied to the tracking task. For this purpose, we evaluate the full system performance with three different detection modules:

\begin{itemize}
\item\textit{Geometric:} is our baseline detection approach which uses the full raw Velodyne information as input. It initially performs a ground floor removal, according to \cite{petrovskaya2009model} and applies a clustering algorithm over the remaining points. Bounding boxes are then extracted as described in Section \ref{sec:bbs}, to obtain the final detections. However, as there is not trustworthy information about the vehicleness of the created clusters, additional geometric constraints are introduced, e.g. no track is created until a corner of the vehicle is identified.

\item\textit{DeepLidar:} is the proposed deep model trained over the full training set. We therefore show the results obtained over the testing dataset, which are provided by the Kitti evaluation server. 

\item\textit{CNN-GT:} its aim is to set the upper bounds of the tracker capacities under ideally lidar vehicle detections. For that, it simulates the perfect output of our convolutional detector by using the ground-truth of our data representation as predictions. As no noise is introduced on the detection step with this approach, the threshold discarding clusters with less than $25$ points is lowered to $4$. It is only evaluated in the training set, as ground-truth is not provided for the test sequences. 
\end{itemize}

The quantitative tracking results of the proposed system are shown in Table \ref{tab:results}. 
Since there is no single ranking criteria to evaluate the tracking task, we follow the Mostly-Tracked (MT), Partly-Tracked (PT) and Mostly-Lost (ML) evaluation measurements from \cite{li2009learning}, as we consider that it reflects better the contributions of the different detector schemas over the final tracking results. This criteria accounts as MT targets those that are successfully tracked for at least the $80\%$ of its life span, whereas as ML the ones with less than $20\%$, and PT the rest.
In addition we include the CLEARMOT \textit{MOTA} metric \cite{bernardin2008evaluating}. It is commonly used due to its expressiveness, as it combines in one single criteria three sources of errors (False Negatives, False Positives and ID-Switches) over the number of ground truth objects. It reports a percentage between $(-\inf,100]$, which takes negative values when the number of errors made by the tracker exceeds the number of total objects in the scene.

\begin{table}[]
\centering
\caption{Quantitative Evaluation on the Kitti Tracking Benchmark}
\label{tab:results}
\begin{tabular}{r||c|c||c|c||c||}
\multicolumn{1}{l||}{} & \multicolumn{2}{c||}{Geometric}                         & \multicolumn{2}{c||}{DeepLidar}                          & \multicolumn{1}{c||}{CNN-GT}                            \\ 
\multicolumn{1}{l||}{} & \multicolumn{1}{c|}{Train} & \multicolumn{1}{c||}{Test} & \multicolumn{1}{c|}{Train} & \multicolumn{1}{c||}{Test} & \multicolumn{1}{c||}{Train} \\ \hline \hline
Mostly Tracked $(\%)$  &   7.4    	&    10.6   	&   -  &    18.5    &    44.5  \\ 
Partly Tracked $(\%)$  &   56.5    	&    45.1   	&   -  &    52.2    &    47.7  \\ 
Mostly Lost  $(\%)$    &   35.9    	&    44.3		&   -  &    29.4    &    7.8  \\ \hline
Recall      $(\%)$     &   46.4   	&    42.1   	&   -  &    55.4    &    79.0  \\ 
Precision   $(\%)$     &   44.1    	&    37.5   	&   -  &    63.8    &    73.9  \\ 
False Alarm Rate 	   &   1.97    	&    2.35   	&   -  &    1.06    &    1.00  \\ \hline
MOTA             	   &  -25.7		&   -38.9   	&   -  &    15.5    &    41.9  \\ \hline

\end{tabular}
\end{table}

The importance of our detector is stated through the noticeable improvements with respect to the geometric baseline method in all the metrics. Our CNN configuration is able to reduce by almost $15\%$ the ML targets, providing better target tracks, which is reflected as an increase of the MT and PT values.
The difference of our \textit{DeepLidar} approach with respect to the ideal \textit{CNN-GT} detector is in fact understandable. Considering that there is no noise introduced by the ideal detector, there is no need for setting a minimum cluster size and therefore farther vehicles can be tracked, which directly reflects in a better MT and MOTA.  
On the other hand, the fact that this ideal system does not achieve perfect results is explained by the own Lidar technology (very far vehicles are impacted by very few points or not even impacted). 

It is noteworthy to mention at this point that the evaluation measurements for the Kitty tracking benchmark are performed on 2D bounding boxes over RGB images. When working only with Lidar information we need to project our 3D tracked bounding boxes to 2D images, which results in lower image pixel-level accuracy and therefore penalise our Lidar-only systems. 
As a fact, Kitti RGB images contains almost $1.4M~(375\!\times1242\!\times~3)$ colour samples. When compared to image tracking methods our CNN inputs only have $64\!\times451\!\times2$, which means that we perform the full vehicle detection and tracking pipeline with just a fraction of $4.13\%$ over the total values of the RGB methods.

%% file: 5_Conclusions_Future.tex
\section{CONCLUSIONS}

In this work we presented a full vehicle detection and tracking system based only on 3D lidar information. It combines a convolutional neural network performing a point-wise vehicle detection, with a multi-object tracker. 
Our CNN-based detector classifies each 3D laser point as belonging to a vehicle or not by using a featured 2D-lidar representation which involves both range and reflectivity information. The resulting positively classified points are then grouped together to create identification hypotheses, and fed to a multi-hypothesis Extended Kalman Filter to track their motion.
We evaluated our system on the KITTI tracking dataset, showing that the inclusion of the CNN-based detection module improves systematically the whole system performance.

%% file: root.bbl
\begin{thebibliography}{10}
\providecommand{\url}[1]{#1}
\csname url@rmstyle\endcsname
\providecommand{\newblock}{\relax}
\providecommand{\bibinfo}[2]{#2}
\providecommand\BIBentrySTDinterwordspacing{\spaceskip=0pt\relax}
\providecommand\BIBentryALTinterwordstretchfactor{4}
\providecommand\BIBentryALTinterwordspacing{\spaceskip=\fontdimen2\font plus
\BIBentryALTinterwordstretchfactor\fontdimen3\font minus
  \fontdimen4\font\relax}
\providecommand\BIBforeignlanguage[2]{{%
\expandafter\ifx\csname l@#1\endcsname\relax
\typeout{** WARNING: IEEEtran.bst: No hyphenation pattern has been}%
\typeout{** loaded for the language `#1'. Using the pattern for}%
\typeout{** the default language instead.}%
\else
\language=\csname l@#1\endcsname
\fi
#2}}

\bibitem{krizhevsky2012imagenet}
A.~Krizhevsky, I.~Sutskever, and G.~E. Hinton, ``Imagenet classification with
  deep convolutional neural networks,'' in \emph{Neural Inf. Process. Syst.
  (NIPS)}, 2012, pp. 1097--1105.

\bibitem{ren2015faster}
S.~Ren, K.~He, R.~Girshick, and J.~Sun, ``Faster {R-CNN}: Towards real-time
  object detection with region proposal networks,'' in \emph{Neural Inf.
  Process. Syst. (NIPS)}, 2015, pp. 91--99.

\bibitem{long2015fully}
J.~Long, E.~Shelhamer, and T.~Darrell, ``Fully convolutional networks for
  semantic segmentation,'' in \emph{IEEE Conf. Comput. Vis. Pattern Recognit.
  (CVPR)}, 2015, pp. 3431--3440.

\bibitem{vvaquero2017flow}
V.~Vaquero, G.~Ros, F.~Moreno-Noguer, A.~M. Lopez, and A.~Sanfeliu, ``Joint
  coarse-and-fine reasoning for deep optical flow.''

\bibitem{Geiger2012CVPR}
A.~Geiger, P.~Lenz, and R.~Urtasun, ``Are we ready for autonomous driving? the
  kitti vision benchmark suite,'' in \emph{IEEE Conf. Comput. Vis. Pattern
  Recognit. (CVPR)}, 2012, pp. 3354--3361.

\bibitem{he2016residual}
K.~He, X.~Zhang, S.~Ren, and J.~Sun, ``Deep residual learning for image
  recognition,'' in \emph{IEEE Conf. Comput. Vis. Pattern Recognit. (CVPR)},
  2016, pp. 770--778.

\bibitem{sermanet2014overfeat}
P.~Sermanet, D.~Eigen, X.~Zhang, M.~Mathieu, R.~Fergus, and Y.~LeCun,
  ``Overfeat: Integrated recognition, localization and detection using
  convolutional networks,'' in \emph{Int. Conf. Learning Representations
  (ICLR)}, 2014.

\bibitem{arras2007using}
K.~O. Arras, O.~M. Mozos, and W.~Burgard, ``Using boosted features for the
  detection of people in {2D} range data,'' in \emph{IEEE Int. Conf. Robotics
  Autom. (ICRA)}, 2007, pp. 3402--3407.

\bibitem{vaquero2016low}
V.~Vaquero, E.~Repiso, A.~Sanfeliu, J.~Vissers, and M.~Kwakkernaat, ``Low cost,
  robust and real time system for detecting and tracking moving objects to
  automate cargo handling in port terminals,'' in \emph{2nd Iberian Robotics
  Conf.}, ser. Adv. Intell. Syst. Comput., vol. 418, 2015, pp. 491--502.

\bibitem{douillard2011segmentation}
B.~Douillard, J.~Underwood, N.~Kuntz, V.~Vlaskine, A.~Quadros, P.~Morton, and
  A.~Frenkel, ``On the segmentation of {3D LIDAR} point clouds,'' in \emph{IEEE
  Int. Conf. Robotics Autom. (ICRA)}, 2011, pp. 2798--2805.

\bibitem{mertz2013moving}
C.~Mertz, L.~E. Navarro-Serment, R.~MacLachlan, P.~Rybski, A.~Steinfeld,
  A.~Suppe, C.~Urmson, N.~Vandapel, M.~Hebert, C.~Thorpe, \emph{et~al.},
  ``Moving object detection with laser scanners,'' \emph{Journal of Field
  Robotics}, vol.~30, no.~1, pp. 17--43, 2013.

\bibitem{mozos2010multi}
O.~M. Mozos, R.~Kurazume, and T.~Hasegawa, ``Multi-part people detection using
  {2D} range data,'' \emph{Intl. Journal of Social Robotics}, vol.~2, no.~1,
  pp. 31--40, 2010.

\bibitem{spinello2010layered}
L.~Spinello, K.~O. Arras, R.~Triebel, and R.~Siegwart, ``A layered approach to
  people detection in {3D} range data,'' in \emph{AAAI Conf. Artif. Intell.
  (AAAI)}, 2010, pp. 1625--1630.

\bibitem{petrovskaya2009model}
A.~Petrovskaya and S.~Thrun, ``Model based vehicle detection and tracking for
  autonomous urban driving,'' \emph{Autonomous Robots}, vol.~26, no. 2-3, pp.
  123--139, 2009.

\bibitem{teichman2011towards}
A.~Teichman, J.~Levinson, and S.~Thrun, ``Towards {3D} object recognition via
  classification of arbitrary object tracks,'' in \emph{IEEE Int. Conf.
  Robotics Autom. (ICRA)}, 2011, pp. 4034--4041.

\bibitem{wang2012could}
D.~Z. Wang, I.~Posner, and P.~Newman, ``What could move? finding cars,
  pedestrians and bicyclists in {3D} laser data,'' in \emph{IEEE Int. Conf.
  Robotics Autom. (ICRA)}, 2012, pp. 4038--4044.

\bibitem{triebel2010segmentation}
R.~Triebel, J.~Shin, and R.~Siegwart, ``Segmentation and unsupervised
  part-based discovery of repetitive objects,'' in \emph{Robotics: Science and
  Systems (RSS)}, 2010, pp. 65--72.

\bibitem{papon2013voxel}
J.~Papon, A.~Abramov, M.~Schoeler, and F.~Worgotter, ``Voxel cloud connectivity
  segmentation-supervoxels for point clouds,'' in \emph{IEEE Conf. Comput. Vis.
  Pattern Recognit. (CVPR)}, 2013, pp. 2027--2034.

\bibitem{wang2015voting}
D.~Z. Wang and I.~Posner, ``Voting for voting in online point cloud object
  detection.'' in \emph{Robotics: Science and Systems (RSS)}, 2015.

\bibitem{behley2012performance}
J.~Behley, V.~Steinhage, and A.~B. Cremers, ``Performance of histogram
  descriptors for the classification of {3D} laser range data in urban
  environments,'' in \emph{IEEE Int. Conf. Robotics Autom. (ICRA)}, 2012, pp.
  4391--4398.

\bibitem{de2013unsupervised}
M.~De~Deuge, A.~Quadros, C.~Hung, and B.~Douillard, ``Unsupervised feature
  learning for classification of outdoor {3D} scans,'' in \emph{Australasian
  Conf. Robotics Automat. (ACRA)}, 2013.

\bibitem{li20163d}
B.~Li, ``{3D} fully convolutional network for vehicle detection in point
  cloud,'' \emph{arXiv preprint arXiv:1611.08069}, 2016.

\bibitem{engelcke2017vote3deep}
M.~Engelcke, D.~Rao, D.~Z. Wang, C.~H. Tong, and I.~Posner, ``Vote3{D}eep: Fast
  object detection in {3D} point clouds using efficient convolutional neural
  networks,'' in \emph{IEEE Int. Conf. Robotics Autom. (ICRA)}, 2017.

\bibitem{jampani2016learning}
V.~Jampani, M.~Kiefel, and P.~V. Gehler, ``Learning sparse high dimensional
  filters: Image filtering, dense {CRF}s and bilateral neural networks,'' in
  \emph{IEEE Conf. Comput. Vis. Pattern Recognit. (CVPR)}, 2016, pp.
  4452--2261.

\bibitem{graham2015sparse}
B.~Graham, ``Sparse {3D} convolutional neural networks,'' in \emph{British
  Machine Vision Conf.}, 2015, pp. 150.1--150.9.

\bibitem{li2016vehicle}
B.~Li, T.~Zhang, and T.~Xia, ``Vehicle detection from {3D} lidar using fully
  convolutional network,'' in \emph{Robotics: Science and Systems (RSS)}, 2016.

\bibitem{chen2016multi}
X.~Chen, H.~Ma, J.~Wan, B.~Li, and T.~Xia, ``Multi-view {3D} object detection
  network for autonomous driving,'' in \emph{IEEE Conf. Comput. Vis. Pattern
  Recognit. (CVPR)}, 2017.

\bibitem{RosArxiv16}
G.~Ros, S.~Stent, P.~F. Alcantarilla, and T.~Watanabe, ``Training constrained
  deconvolutional networks for road scene semantic segmentation,'' \emph{{arXiv
  preprint abs/1604.01545}}, 2016.

\bibitem{he2015delving}
K.~He, X.~Zhang, S.~Ren, and J.~Sun, ``Delving deep into rectifiers: Surpassing
  human-level performance on imagenet classification,'' in \emph{The IEEE
  International Conference on Computer Vision (ICCV)}, 2015.

\bibitem{li2009learning}
Y.~Li, C.~Huang, and R.~Nevatia, ``Learning to associate: Hybridboosted
  multi-target tracker for crowded scene,'' in \emph{Computer Vision and
  Pattern Recognition, 2009. CVPR 2009. IEEE Conference on}.\hskip 1em plus
  0.5em minus 0.4em\relax IEEE, 2009, pp. 2953--2960.

\bibitem{bernardin2008evaluating}
K.~Bernardin and R.~Stiefelhagen, ``Evaluating multiple object tracking
  performance: the clear mot metrics,'' \emph{EURASIP Journal on Image and
  Video Processing}, vol. 2008, no.~1, pp. 1--10, 2008.

\end{thebibliography}
